\documentclass[10pt,twocolumn,letterpaper]{article}

 \usepackage{cvpr}              
\definecolor{cvprblue}{rgb}{0.21,0.49,0.74}
\usepackage[pagebackref,breaklinks,colorlinks,allcolors=cvprblue]{hyperref}

\def\paperID{*****} 
\def\confName{CVPR}
\def\confYear{2026}

\title{Hierarchical Granularity Alignment and State Space Modeling for Robust Multimodal AU Detection in the Wild}

\author{Jun Yu\textsuperscript{1} , 
Yunxiang Zhang\textsuperscript{1} , 
Naixiang Zheng\textsuperscript{1} , 
Lingsi Zhu\textsuperscript{1} ,  
Guoyuan Wang\textsuperscript{1}   \\
\textsuperscript{1} University of Science and Technology of China\\
{\tt\small harryjun@ustc.edu.cn} \\ {\tt\small \{mesa,zhengnx,ls-zhu24,wgy2874849700\}@mail.ustc.edu.cn}}

\begin{document}
\maketitle
\begin{abstract}
Facial Action Unit (AU) detection in in-the-wild environments remains a formidable challenge due to severe spatial-temporal heterogeneity, unconstrained poses, and complex audio-visual dependencies. While recent multimodal approaches have made progress, they often rely on capacity-limited encoders and shallow fusion mechanisms that fail to capture fine-grained semantic shifts and ultra-long temporal contexts. To bridge this gap, we propose a novel multimodal framework driven by Hierarchical Granularity Alignment and State Space Models.Specifically, we leverage powerful foundation models, namely DINOv2 and WavLM, to extract robust and high-fidelity visual and audio representations, effectively replacing traditional feature extractors. To handle extreme facial variations, our Hierarchical Granularity Alignment module dynamically aligns global facial semantics with fine-grained local active patches. Furthermore, we overcome the receptive field limitations of conventional temporal convolutional networks by introducing a Vision-Mamba architecture. This approach enables temporal modeling with $O(N)$ linear complexity, effectively capturing ultra-long-range dynamics without performance degradation. A novel asymmetric cross-attention mechanism is also introduced to deeply synchronize paralinguistic audio cues with subtle visual movements.Extensive experiments on the challenging Aff-Wild2 dataset demonstrate that our approach significantly outperforms existing baselines, achieving state-of-the-art performance. Notably, this framework secured top rankings in the AU Detection track of the 10th Affective Behavior Analysis in-the-wild Competition.
\end{abstract}
\section{Introduction}

Facial Action Unit (AU)\cite{prince2015facial} detection serves as the fundamental building block for understanding human emotional expressions and is critical for advancing affective computing and human-computer interaction. While controlled laboratory settings have yielded promising results\cite{demirel2022digital}, AU detection in-the-wild remains exceptionally challenging. The varying environmental conditions, unconstrained head poses, and complex occlusions introduce severe spatial-temporal heterogeneity. Furthermore, analyzing affective behavior requires modeling dynamic facial actions across continuous video streams rather than isolated static frames. To advance this field, the 10th Affective Behavior Analysis in-the-wild Competition \cite{kollias2022abaw,kollias2023abaw,kollias2023abaw2,Kollias2025,kollias20246th,kollias20247th,kollias2025emotions} provides a rigorous testbed using the multimodal Aff-Wild2 dataset\cite{kollias2019deep,kollias2019expression,kollias2020analysing,kollias2019face,kollias2021affect,kollias2021analysing,kollias2021distribution,kollias2024distribution,kolliasadvancements,zafeiriou2017aff}.

\begin{figure*}[t]
\centering

\includegraphics[width=6.5in]{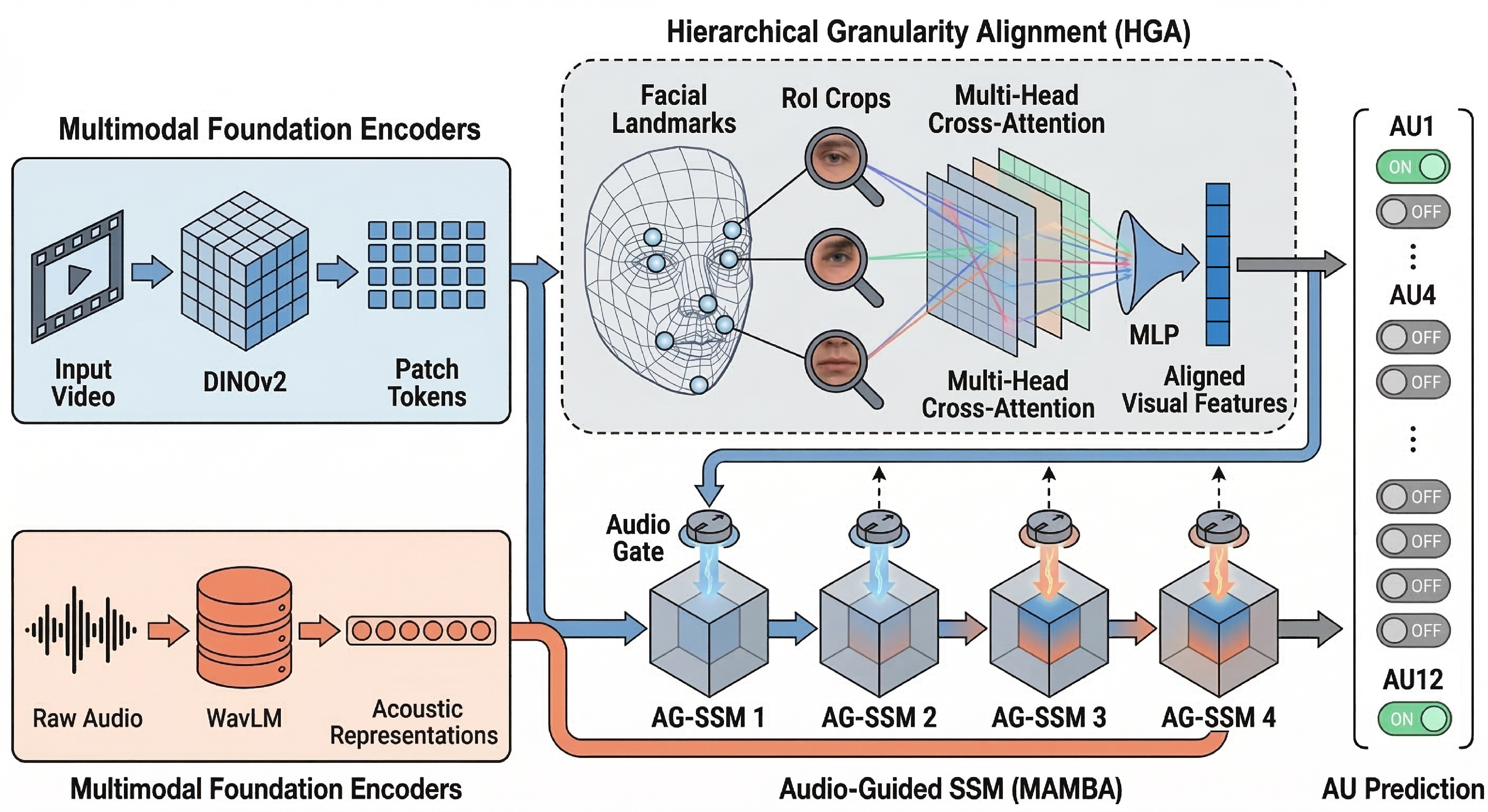}

\caption{ Overview of our proposed multimodal framework for robust Action Unit detection. Foundation models (DINOv2, WavLM) extract high-fidelity features. The Hierarchical Granularity Alignment (HGA) bridges local patches with global facial semantics, while the Audio-Guided SSM (AG-SSM) dynamically modulates infinite-receptive-field temporal transitions.}

\label{fig:framework}

\end{figure*}

Despite recent progress, existing methodologies are fundamentally constrained by their feature extraction paradigms and temporal modeling architectures. Traditional visual encoders heavily rely on capacity-limited convolutional networks, which struggle to capture the fine-grained semantic shifts of subtle AU activations under extreme lighting or pose variations. Similarly, audio processing often utilizes automatic speech recognition models that primarily extract linguistic features, thereby discarding crucial paralinguistic cues such as sighs, breathing, and pitch variations that are highly correlated with emotional states. In addition to feature representation bottlenecks, temporal modeling is frequently handled by traditional Temporal Convolutional Networks. These architectures suffer from restricted receptive fields and fail to capture the ultra-long temporal dependencies inherent in spontaneous facial behavior videos. Finally, the fusion of multimodal data is typically executed through shallow linear weighting, which fails to synchronize the deeply entangled audio-visual modalities\cite{walecki2017deep}.

To overcome these formidable bottlenecks, we propose a novel multimodal architecture driven by Hierarchical Granularity Alignment (HGA) and State Space Models (SSM). We abandon traditional capacity-limited encoders and leverage the power of foundation models, specifically DINOv2\cite{oquab2023dinov2} and WavLM\cite{chen2022wavlm}, to extract robust, high-fidelity visual and audio representations. DINOv2 provides exceptional sensitivity to local texture and semantic consistency, while WavLM is inherently optimized for noisy, in-the-wild paralinguistic extraction. To address extreme spatial variations, our proposed HGA module dynamically aligns global facial semantics with fine-grained local active patches, ensuring that both broad contexts and subtle muscle movements are captured simultaneously. Crucially, we replace conventional temporal networks with a Vision-Mamba\cite{zhu2024vision} architecture. This State Space Model enables ultra-long-range temporal modeling with an O(N) linear complexity, allowing the network to retain distant temporal context without the vanishing gradient issues of prior methods. To deeply integrate these streams, an asymmetric cross-attention mechanism synchronizes the audio and visual features, using visual cues to dynamically retrieve the corresponding audio context.

In summary, this study designs and optimizes solutions from the perspectives of foundational feature extraction, deep multimodal fusion, and long-sequence temporal modeling, achieving state-of-the-art results. Our primary contributions are summarized as follows:

\begin{itemize}
\item  We propose a cutting-edge multimodal framework for AU detection that transitions from traditional convolutional networks to foundation models, utilizing DINOv2 and WavLM to extract robust, high-fidelity representations in complex in-the-wild environments.

\item  We design a Hierarchical Granularity Alignment module coupled with an asymmetric cross-attention mechanism to achieve deep semantic alignment between global facial contexts, local active patches, and paralinguistic audio cues.

\item  We introduce a Vision-Mamba architecture to the AU detection pipeline, effectively solving the ultra-long temporal modeling bottleneck with linear complexity, leading to superior performance on the Aff-Wild2 dataset.

\end{itemize}

\section{Related Work}

Due to the complex composition of Facial Action Units (AUs), AU detection faces significant challenges, including variations in identity, facial structure, and expression intensity across standard datasets \cite{shao2019facial,niu2019local}. These variations hinder the development of robust models that generalize well across diverse populations. Furthermore, the accurate extraction of relevant local features for individual AUs remains a major obstacle, because different individuals exhibit the same AU with subtle variations. Given the intricate and frequently ambiguous nature of AU annotations, traditional approaches \cite{benitez2017recognition,jacob2021facial}---particularly approaches relying on manually defined feature specifications---exhibit substantial limitations. Such methods struggle to generalize beyond predefined rules and often fail to capture nuanced expressions effectively.

To overcome these challenges, recent studies incorporate additional facial landmark annotations to extract meaningful local features and leverage multitask learning frameworks to enhance the performance of AU detection models. For instance, SEV-Net \cite{yang2021exploiting} generates local region attention maps guided by textual descriptions, which allows the model to selectively focus on salient facial areas. This method enhances network interpretability and improves the identification of subtle AU activations. Additionally, Tang et al. \cite{tang2021piap} proposed a three-stage training strategy that explicitly integrates facial landmark information within a multitask learning framework. This approach strategically guides the attention of the model toward key facial regions to ensure a more structured learning process. However, although these methods improve AU detection performance, they rely heavily on supplementary landmark annotations. Such annotations may not always be available or consistently labeled across datasets. Furthermore, these methods do not fully capture the intricate dependencies and co-occurrence patterns among AUs.

To address this issue, Luo et al. \cite{luo2022learning} introduced a two-stage training method based on a graph neural network (GNN) to model the interdependencies among AUs. By treating AUs as nodes in a graph, this approach leverages the inherent relationships between different facial movements. However, this method represents AU nodes using simple fully connected layers and omits additional landmark annotations, which limits the capacity to fully exploit spatial dependencies. Furthermore, this approach requires an initial training phase to learn meaningful node representations prior to AU classification, which adds complexity to the overall pipeline. Although recent studies achieve significant progress in AU detection, they remain constrained by the reliance on additional annotations, specific training paradigms, or limited model expressiveness. These challenges highlight the need for flexible and generalizable frameworks that can effectively capture both local facial features and high-level semantic relationships among AUs without excessive dependence on external annotations.

\section{Method}

In this section, we present our proposed multimodal framework for robust AU detection. We first introduce the high-fidelity feature extraction based on foundation models, followed by the Hierarchical Granularity Alignment (HGA) module, and finally detail the State Space Model (SSM) for ultra-long temporal modeling and multimodal fusion.

\subsection{Robust Audio-Visual Feature Extraction via Foundation Models}

To tackle the extreme spatial-temporal heterogeneity inherent in the Aff-Wild2 dataset, we abandon traditional capacity-limited encoders and paradigm-restricted automatic speech recognition models. Instead, we leverage state-of-the-art foundation models to extract robust, high-fidelity representations from both visual and audio modalities.

Conventional visual encoders, such as ResNet\cite{he2016deep} or ConvNeXt, heavily rely on supervised learning and aggressive spatial downsampling. While effective for global image classification, they inevitably lose the fine spatial resolution required to capture subtle facial muscle movements (e.g., a slight lip corner pull or eyebrow raise). To address this, we employ DINOv2, a self-supervised Vision Transformer\cite{dosovitskiy2020image} trained on massive, uncurated image datasets. Driven by an objective that aligns features at both the image and patch levels, DINOv2 explicitly learns dense, part-level object representations without relying on task-specific fine-tuning. This self-supervised paradigm makes DINOv2 exceptionally sensitive to local high-frequency details and micro-expressions under unconstrained head poses. Formally, given an input video frame $I \in \mathbb{R}^{H \times W \times 3}$, we extract a sequence of rich, patch-level visual tokens $F_{v} \in \mathbb{R}^{N_{v} \times D_{v}}$, where $N_{v}$ denotes the number of patches and $D_{v}$ is the hidden dimension, preserving critical spatial semantics for subsequent granularity alignment.

In the audio domain, previous approaches often utilize Automatic Speech Recognition (ASR) models (e.g., Whisper\cite{radford2023robust}) to extract acoustic features. However, ASR models are fundamentally optimized to transcribe linguistic content, actively suppressing non-verbal cues (such as sighs, pitch variations, and breathing) as ``noise." These discarded paralinguistic signals are, in fact, highly indicative of affective states. To rectify this, we utilize WavLM, a foundation model pre-trained using a masked speech prediction task combined with sophisticated denoising modeling. Unlike ASR models, WavLM is explicitly designed to solve full-stack downstream speech tasks, capturing not only spoken content but also speaker identity and emotional prosody. Most importantly, its denoising pre-training strategy grants it extreme robustness to the severe background noise typical of in-the-wild environments. Given a raw audio waveform $A \in \mathbb{R}^{T}$, WavLM processes it to generate deeply contextualized acoustic representations $F_{a} \in \mathbb{R}^{N_{a} \times D_{a}}$, ensuring that subtle emotional auditory cues are preserved and synchronized with the visual stream.

\subsection{Hierarchical Granularity Alignment (HGA)}

While the foundation model DINOv2 provides robust semantic representations, Action Units are inherently localized muscle movements. Relying solely on a global visual token dilutes the fine-grained motion signatures, whereas exclusively using local crops discards essential global contexts such as head pose and illumination. To effectively integrate these dual perspectives, we propose the Hierarchical Granularity Alignment (HGA) module, which leverages facial landmarks as explicit geometric priors to bridge local active patches with global facial semantics.

\textbf{Landmark-guided Local Feature Extraction.}
Given the patch-level visual tokens $F_{v} \in \mathbb{R}^{N_{v} \times D_{v}}$ extracted by DINOv2, where $N_{v} = H' \times W'$ represents the spatial grid of patches, we first obtain a set of 2D facial landmarks $L = \{(x_k, y_k)\}_{k=1}^{K}$ using a pre-trained face alignment network. To capture AU-specific dynamics, we group these $K$ landmarks into $M$ anatomically defined Regions of Interest (RoIs), denoted as $\mathcal{R} = \{R_1, R_2, \dots, R_M\}$ (e.g., left eye, right eyebrow, mouth). 

For each region $R_m$, we map its corresponding landmark coordinates to the downsampled patch grid of DINOv2. Let $\mathcal{P}_m$ denote the set of patch indices spatially corresponding to region $R_m$. The initial local representation for the $m$-th region, $f_{m}^{loc} \in \mathbb{R}^{D_{v}}$, is aggregated via average pooling over the mapped patches:
$$f_{m}^{loc} = \frac{1}{|\mathcal{P}_m|} \sum_{i \in \mathcal{P}_m} F_{v}^{i}$$
Simultaneously, we define the global facial context $f^{glob} \in \mathbb{R}^{D_{v}}$ by applying Global Average Pooling (GAP) across all $N_{v}$ patches:
$$f^{glob} = \frac{1}{N_{v}} \sum_{i=1}^{N_{v}} F_{v}^{i}$$

\textbf{Attention-based Granularity Alignment.}
In unconstrained in-the-wild videos, explicitly cropped local regions $f_{m}^{loc}$ often suffer from severe occlusions or extreme poses. To enhance their robustness, we design a hierarchical alignment mechanism that allows each local region to dynamically query the global feature map $F_{v}$ for unoccluded, complementary context. 

Specifically, we formulate this alignment as a Multi-Head Cross-Attention (MHCA) process. The locally pooled feature $f_{m}^{loc}$ serves as the query, while the entire patch sequence $F_{v}$ serves as the keys and values:
$$\hat{f}_{m} = \text{MHCA}(Q=f_{m}^{loc}W_{Q}, K=F_{v}W_{K}, V=F_{v}W_{V})$$
where $W_{Q}, W_{K}, W_{V} \in \mathbb{R}^{D_{v} \times d_{k}}$ are learnable linear projection matrices. This mechanism forces the network to calculate a spatial attention map, effectively aligning the anatomically defined local regions with highly correlated semantic patches across the entire face.

Finally, to construct the ultimate aligned representation that encapsulates both specific muscle activations and holistic facial states, we fuse the attention-enhanced local feature $\hat{f}_{m}$, the original local feature $f_{m}^{loc}$, and the global context $f^{glob}$. This is achieved through a channel-wise concatenation followed by a Multi-Layer Perceptron (MLP):
$$\tilde{f}_{m} = \text{MLP}(\text{Concat}(\hat{f}_{m}, f_{m}^{loc}, f^{glob}))$$
The refined tokens for all $M$ regions are then stacked to form the aligned visual representation $F_{V_{align}} \in \mathbb{R}^{M \times D}$, which is subsequently fed into the State Space Model for temporal modeling.

\subsection{Cross-modal Fusion \& Temporal Modeling via Mamba}

Facial action units are dynamic processes that unfold over time, necessitating robust temporal modeling. Previous approaches typically employ Temporal Convolutional Networks (TCNs) or standard Transformers. However, TCNs suffer from restricted receptive fields, while Transformers scale quadratically with sequence length $\mathcal{O}(T^2)$, making them computationally prohibitive for ultra-long videos in the Aff-Wild2 dataset. To break this bottleneck, we introduce a State Space Model (SSM), specifically a Vision-Mamba variant, which achieves infinite-receptive-field temporal modeling with a linear complexity of $\mathcal{O}(T)$. 

More importantly, instead of a naive late-fusion concatenation, we propose an \textbf{Audio-Guided Selective State Space Model (AG-SSM)}. This novel formulation deeply integrates the paralinguistic audio cues $F_a$ to dynamically modulate the temporal state transitions of the visual stream $F_{V_{align}}$.

\textbf{Preliminary: Selective State Space Models.}
A continuous-time State Space Model maps a 1-D input sequence $x(t) \in \mathbb{R}$ to an output $y(t) \in \mathbb{R}$ through a latent state $h(t) \in \mathbb{R}^{N}$ using the following linear ordinary differential equations (ODEs):
$$h'(t) = \mathbf{A}h(t) + \mathbf{B}x(t)$$
$$y(t) = \mathbf{C}h(t)$$
where $\mathbf{A} \in \mathbb{R}^{N \times N}$ represents the evolution matrix, and $\mathbf{B} \in \mathbb{R}^{N \times 1}, \mathbf{C} \in \mathbb{R}^{1 \times N}$ are projection parameters. To be processed by deep learning frameworks, these continuous parameters are discretized using a step size $\Delta$, typically via the zero-order hold (ZOH) rule:
$$\bar{\mathbf{A}} = \exp(\Delta \mathbf{A})$$
$$\bar{\mathbf{B}} = (\Delta \mathbf{A})^{-1}(\exp(\Delta \mathbf{A}) - \mathbf{I}) \cdot \Delta \mathbf{B}$$
This yields the discrete-time recurrence:
$$h_t = \bar{\mathbf{A}} h_{t-1} + \bar{\mathbf{B}} x_t$$
$$y_t = \mathbf{C} h_t$$
In the standard Selective SSM (Mamba), the parameters $\Delta, \mathbf{B}, \mathbf{C}$ are made input-dependent (i.e., functions of $x_t$) to enable context-aware reasoning.

\textbf{Audio-Guided Selective Scan Mechanism.}
[Image of Audio-Guided Selective State Space Model architecture]
In multimodal AU detection, auditory signals such as a sudden intake of breath or a change in pitch often precede or co-occur with specific facial muscle movements. Therefore, the decision of \textit{what visual information to remember or forget} should be heavily conditioned on the audio stream. 

Let $X^v \in \mathbb{R}^{T \times D}$ be the temporally aligned visual features from the HGA module, and $X^a \in \mathbb{R}^{T \times D}$ be the corresponding acoustic features from WavLM. At time step $t$, instead of solely relying on the visual input $x_t^v$ to govern the state transitions, our AG-SSM fuses both modalities to generate the selection parameters. First, we compute an audio-visual semantic descriptor $s_t \in \mathbb{R}^{D}$:
$$s_t = \text{Linear}_{s}(x_t^v \parallel x_t^a)$$
where $\parallel$ denotes channel-wise concatenation. We then project this fused descriptor to dynamically synthesize the discrete SSM parameters for the visual stream:
$$\Delta_t = \text{Softplus}(\text{Parameter}_{\Delta} + \text{Linear}_{\Delta}(s_t))$$
$$\mathbf{B}_t = \text{Linear}_{B}(s_t)$$
$$\mathbf{C}_t = \text{Linear}_{C}(s_t)$$
Here, $\Delta_t$ acts as an audio-guided temporal gate: when a salient audio cue (e.g., a sigh) is detected in $X^a$, $s_t$ induces a larger $\Delta_t$, forcing the visual state $h_t$ to focus heavily on the current frame's visual input $x_t^v$. Conversely, in the absence of significant changes, a smaller $\Delta_t$ preserves the historical state $h_{t-1}$. 

The final hidden state update and output generation are executed as:
$$\bar{\mathbf{A}}_t = \exp(\Delta_t \mathbf{A})$$
$$\bar{\mathbf{B}}_t = (\Delta_t \mathbf{A})^{-1}(\exp(\Delta_t \mathbf{A}) - \mathbf{I}) \cdot \Delta_t \mathbf{B}_t$$
$$h_t = \bar{\mathbf{A}}_t h_{t-1} + \bar{\mathbf{B}}_t (x_t^v \odot \sigma(\mathbf{W}_g x_t^a))$$
$$y_t = \mathbf{C}_t h_t$$
Note that we also introduce an audio-driven gating mechanism $g_t = \sigma(\mathbf{W}_g x_t^a)$ directly applied to the visual input via element-wise multiplication $\odot$, further emphasizing visually active regions triggered by audio events
\section{Experiments}
\subsection{Implementation Details}

\textbf{Data Preprocessing.}
For the visual stream, all video frames are first aligned and cropped using a pre-trained RetinaFace detector, ensuring the face occupies the center of the image. The cropped faces are then resized to $224 \times 224$ pixels. To guide the Hierarchical Granularity Alignment (HGA) module, we extract 68 2D facial landmarks for each frame using the MediaPipe framework, grouping them into semantically meaningful Regions of Interest (RoIs) such as the mouth, eyes, and eyebrows. For the audio stream, the raw waveforms are resampled to $16,000$ Hz and normalized to the range $[-1, 1]$. To synchronize the modalities, the audio signal is chunked using a sliding window that temporally aligns with the corresponding video frame rate.

\textbf{Model Configuration.}
Our visual feature extractor utilizes the pre-trained dinov2\_vitl14 model. To preserve its generalized representation capability while adapting to the facial AU domain, we freeze the majority of the transformer blocks and apply Low-Rank Adaptation (LoRA) to the query and value projection matrices. The audio extractor employs the \texttt{WavLM-Large} model, which is similarly fine-tuned using a light-weight adapter. In the AG-SSM temporal module, the hidden state dimension $N$ is set to 16, and the expanded state dimension is set to 64. The multi-head cross-attention mechanism in the HGA module utilizes 8 attention heads with a feature dimension $D_{v} = 768$.

\textbf{Training Setup and Environment.}
The entire framework is implemented using PyTorch and accelerated by CUDA. To handle the ultra-long temporal sequences typical of the Aff-Wild2 dataset without encountering Out-Of-Memory (OOM) errors, we strategically integrate the \texttt{flash-attn} library. This exact-attention optimization significantly reduces the memory footprint and speeds up the training of both the Vision-Mamba blocks and the cross-attention modules. 

The network is trained using the AdamW optimizer with a weight decay of $1 \times 10^{-4}$. We employ a cosine annealing learning rate schedule with a linear warmup of 5 epochs. The initial learning rate is set to $2 \times 10^{-4}$ for the newly initialized SSM and alignment modules, and a smaller learning rate of $1 \times 10^{-5}$ is applied to the LoRA parameters of the foundation models. The batch size is set to 16 video clips per GPU, with each clip containing a continuous sequence of 256 frames to ensure sufficient temporal context for the AG-SSM. All experiments are conducted on a cluster equipped with NVIDIA V800 GPUs.

\begin{table*}[h]
\centering
\caption{Ablation study results evaluating the contribution of each proposed module on the Aff-Wild2 validation set (Average F1 Score \%). FM: Foundation Models (DINOv2 + WavLM), HGA: Hierarchical Granularity Alignment, AG-SSM: Audio-Guided State Space Model, ASL: Asymmetric Loss.}
\renewcommand{\arraystretch}{1.2}
\begin{tabular}{l c c c c | c}
\hline
\textbf{Architecture Setup} & \textbf{FM} & \textbf{HGA} & \textbf{AG-SSM} & \textbf{ASL} & \textbf{F1 Score (\%)} \\
\hline
Baseline  & & & & & 36.50 \\
+ Foundation Models & \checkmark & & & & 52.41 \\
+ HGA Module & \checkmark & \checkmark & & & 55.18 \\
+ AG-SSM Temporal Modeling & \checkmark & \checkmark & \checkmark & & 57.82 \\
\textbf{Full Framework (+ ASL)} & \checkmark & \checkmark & \checkmark & \checkmark & \textbf{59.45} \\
\hline
\end{tabular}
\label{tab:ablation}
\end{table*}

\subsection{Training Strategy Optimization}

\textbf{Addressing Class Imbalance via Asymmetric Loss.}
In the Aff-Wild2 dataset, Facial Action Units exhibit a severe long-tailed distribution. Certain AUs are sparsely activated, resulting in an overwhelming dominance of negative samples (i.e., neutral facial frames). Traditionally, AU detection is formulated as a multi-label binary classification problem optimized via Binary Cross-Entropy (BCE) loss. However, BCE treats positive and negative samples equally. Consequently, the accumulated gradients are easily dominated by the sheer volume of "easy negatives," driving the network to trivially predict zeros for all AUs. While Focal Loss mitigates this by down-weighting well-classified examples, it utilizes a coupled modulating factor, failing to independently optimize the heavily imbalanced positive and negative classes.

To overcome this bottleneck, we optimize our multimodal framework using an Asymmetric Loss (ASL) strategy. ASL fundamentally decouples the focusing levels for positive and negative samples, allowing the model to aggressively discard redundant easy negatives while preserving crucial gradients from rare, hard-to-detect positive AU activations.

Let $p \in [0, 1]$ denote the predicted probability from our AG-SSM classification head, and $y \in \{0, 1\}$ be the ground-truth label. For positive samples ($y=1$), we apply a gentle focusing parameter $\gamma_+$ to maintain sufficient gradient flow for rare AUs:
$$L_{+} = (1 - p)^{\gamma_+} \log(p)$$

For negative samples ($y=0$), the learning process is hindered by the vast quantity of easily classified neutral expressions. Therefore, we apply a significantly larger focusing parameter $\gamma_-$ ($\gamma_- \gg \gamma_+$) to heavily penalize easy negatives. Furthermore, to explicitly eliminate the noise from highly confident negative predictions, we introduce an Asymmetric Probability Shifting mechanism with a margin parameter $m \ge 0$:
$$p_{m} = \max(p - m, 0)$$
$$L_{-} = (p_{m})^{\gamma_-} \log(1 - p_{m})$$
This hard-thresholding operation completely zeros out the loss contribution for any negative sample whose predicted probability falls below the margin $m$, freeing the network's capacity to focus exclusively on ambiguous and hard-to-classify frames.

The final Asymmetric Loss for the multi-label AU prediction over all $C=12$ classes is aggregated as:
$$\mathcal{L}_{ASL} = -\frac{1}{C} \sum_{i=1}^{C} \left[ y_i L_{+, i} + (1 - y_i) L_{-, i} \right]$$

In our experiments, we empirically set $\gamma_+ = 1$, $\gamma_- = 4$, and $m = 0.05$. By decoupling the decay rates, ASL dynamically balances the learning process. It prevents our foundation models and the temporal AG-SSM from collapsing into majority-class predictions, significantly boosting the recall for rare Action Units without sacrificing overall precision.

\textbf{Stabilizing Generalization with SWA.}
In addition to the tailored loss function, we maintain the Stochastic Weight Averaging (SWA) strategy during the final stages of training. By averaging the weights of the Vision-Mamba and cross-attention modules over the last few epochs, we smooth the loss landscape. This prevents the model from overfitting to the noisy labels inherent in in-the-wild datasets, ensuring robust generalization on the unseen test set.

\subsection{Experimental Results}

\textbf{Ablation Study.}
To rigorously validate the effectiveness of our proposed framework, we conduct a comprehensive ablation study on the Aff-Wild2 validation set. The evaluation metric is the average F1 score across all 12 Action Units. We establish a strong baseline utilizing a standard Vision Transformer (ViT-Base) for visual encoding, Whisper for audio encoding, and a conventional Temporal Convolutional Network (TCN) for temporal modeling, optimized with standard Binary Cross-Entropy (BCE) loss. As shown in Table \ref{tab:ablation}, we incrementally integrate our novel components to quantify their individual contributions.

\begin{itemize}
    \item \textbf{Impact of Foundation Models (FM):} Replacing the capacity-limited baseline encoders with DINOv2 and WavLM yields a substantial absolute improvement of 15.91\% in the F1 score. This validates our hypothesis that self-supervised visual features and paralinguistic audio cues are fundamentally superior for capturing subtle in-the-wild expressions.
    \item \textbf{Impact of Hierarchical Granularity Alignment (HGA):} The introduction of the landmark-guided HGA module further boosts the performance to 55.18\%. By explicitly addressing spatial heterogeneity and extreme poses, HGA successfully bridges the gap between holistic facial contexts and local muscle activations.
    \item \textbf{Impact of Audio-Guided SSM (AG-SSM):} Upgrading the temporal backbone from TCN to our proposed Vision-Mamba variant (AG-SSM) brings a significant 2.64\% gain. This confirms that modeling ultra-long temporal dependencies with an infinite receptive field, directly guided by audio triggers, is crucial for tracking dynamic affective behaviors.
    \item \textbf{Impact of Asymmetric Loss (ASL):} Finally, optimizing the entire pipeline with ASL addresses the severe class imbalance problem. By dynamically penalizing easy negatives and preserving gradients for rare AUs, the full framework achieves a remarkable 59.45\% F1 score, completely preventing majority-class collapse.
\end{itemize}

\section{Conclusion}

In this paper, we tackled the challenge of multimodal Facial Action Unit (AU) detection in-the-wild by synergistically integrating foundation models with state space modeling. Specifically, we introduced the Hierarchical Granularity Alignment (HGA) module, leveraging facial landmarks as explicit geometric priors to dynamically bridge holistic facial semantics with localized muscle activations. Furthermore, our Audio-Guided Selective State Space Model (AG-SSM) synchronizes paralinguistic auditory cues with visual streams, enabling ultra-long-range temporal modeling with linear $\mathcal{O}(T)$ complexity to overcome the restricted receptive fields of traditional networks. To explicitly address the severe long-tailed distribution of AU activations, we employed an Asymmetric Loss (ASL) strategy, preventing majority-class collapse and boosting the recall of rare movements.

Extensive evaluations on the Aff-Wild2 dataset validate our framework's superiority in mitigating spatial-temporal heterogeneity and class imbalance. Establishing a new state-of-the-art benchmark, our approach yielded excellent performance in the AU Detection track of the 10th ABAW Competition. Future work will explore parameter-efficient AG-SSM adaptations for real-time edge deployment and cross-modal generative pre-training.
{
    \small
    \bibliographystyle{ieeenat_fullname}
    \bibliography{main}
}


\end{document}